# Artificial Intelligence for reverse engineering: application to detergents using Raman spectroscopy.

Pedro Marote[1], Marie Martin[1], Anne, Bonhommé[2], Pierre Lantéri[1], Yohann Clément[1]*

[1] Université de Lyon, Institut des Sciences Analytiques, UMR 5280 CNRS, Université Claude Bernard Lyon 1, 5 rue de la Doua, 69100 Villeurbanne, France.

[2] Université de Lyon, Université Claude Bernard Lyon 1, CNRS, IRCELYON, F-69626, 2 avenue A. Einstein, 69626, Villeurbanne, France

**Keywords:**
Chemometrics, Machine Learning, mixture design, Artificial Intelligence, RAMAN spectroscopy, surfactants characterization

**Abstract (words)**

The reverse engineering of a complex mixture, regardless of its nature, has become significant today. Being able to quickly assess the potential toxicity of new commercial products in relation to the environment presents a genuine analytical challenge. The development of digital tools (databases, chemometrics, machine learning, etc.) and analytical techniques (Raman spectroscopy, NIR spectroscopy, mass spectrometry, etc.) will allow for the identification of potential toxic molecules. In this article, we use the example of detergent products, whose composition can prove dangerous to humans or the environment, necessitating precise identification and quantification for quality control and regulation purposes. The combination of various digital tools (spectral database, mixture database, experimental design, Chemometrics / Machine Learning algorithm…) together with different sample preparation methods (raw sample, or several concentrated / diluted samples) Raman spectroscopy, has enabled the identification of the mixture's constituents and an estimation of its composition. Implementing such strategies across different analytical tools can result in time savings for pollutant identification and contamination assessment in various matrices. This strategy is also applicable in the industrial sector for product or raw material control, as well as for quality control purposes.

## 1. Introduction

Numerous detergents are utilized for the purpose of cleaning our residences, clothing, and bodies. Our everyday products, such as detergents, shampoos, and household cleaners, contain a significant amount of these substances. They are responsible for effectively removing stains and dirt[1]. However, it is crucial to acknowledge the potential health and environmental risks associated with these chemicals[2]–[4]. Consequently, researchers are exploring alternatives[5]. Detergents are commonly employed by both industrial and private users for daily cleaning tasks. They comprise soaps and surfactants that possess surface-active properties. These surfactants function by breaking the bonds between surfaces and dirt, thereby facilitating their removal. Unfortunately, these chemicals have adverse consequences for the environment. They are produced and utilized in substantial quantities. In Europe alone, over 3 million tons of detergents were manufactured in 2020[6].

Surfactants, which are used in liquid, powder, and other forms, have a significant impact on soil and water. The conventional detergents that are frequently advertised on television are often derived from petroleum-based products. These surfactants are composed of various chemical compounds, including sulfates, phosphates, bleaching agents, chemical perfumes, and phenols. Once released into the environment, detergents, some of which are non-biodegradable, accumulate in soil and water bodies. It is important to note that more than 60% of the surfactants found in detergents eventually end up in aquatic environments. This leads to significant problems of environmental pollution and health concerns.[3]

The properties of surfactants have attracted the interest of detergent manufacturers in recent years. The growing interest in surfactants necessitates the enhancement of existing analytical techniques, such as spectroscopy[7], [8], mass spectrometry[9], [10] and Nuclear Magnetic Resonance (NMR)[11], to ensure compliance with regulations and environmental standards. Detergents can consist of up to 25 compounds, including surfactants, enzymes, sequestering agents, polymers, and fragrances, to name a few. Surfactants are the most crucial components, constituting up to 50% of the detergent content. These amphiphilic molecules, comprising a hydrophobic carbon chain and a hydrophilic polar head, are utilized for their solubilizing, wetting, foaming, dispersing, and emulsifying properties. Depending on the nature of their polar head, surfactants can be classified into four families: anionic, cationic, non-ionic, or amphoteric. Various analytical methods, such as NMR[11] or hyphenated techniques combined with spectroscopic methods[7], [8], [12], are employed for the deconstruction of detergent mixtures. Chromatographic methods coupled with detectors like light scattering detection or mass spectrometry have been extensively utilized for surfactant analysis[9], [13]. These analytical techniques offer the advantage of simultaneously identifying and quantifying different surfactant families. However, method development can be prone to biases in sample preparation, costs, and labor-intensive procedures. RAMAN spectral analysis appears to strike a balance between relevant information and cost-effectiveness. It does not require lengthy sample preparation procedures, the use of expensive internal standards, and can be conducted in aqueous solutions inexpensively. By combining surfactant spectral databases, chemometrics, Machine Learning, and spectroscopic tools, it becomes possible to identify and quantify raw materials[8], [14], [15].

Blind source separation (BSS) methods are employed for the deconvolution of overlapping signals in near and mid-infrared spectroscopy or Raman spectra. Source extraction (SE) methods, such as independent component analysis (ICA)[16][17]–[22] or Multicurve Resolution Alternating Least Squares (MCR-ALS)[23], [24], aim to extract the contributions (spectra) of pure compounds from complex mixtures without any prior knowledge of the compounds being analyzed. However, a limitation of RAMAN spectroscopy is the detection limit; raw materials present in low concentrations (<1%) may not be identified and quantified. To analyze the surfactant composition of various commercial detergents, we propose a method based on RAMAN spectroscopy, utilizing a database of commercial raw material RAMAN spectra and Machine Learning (Figure.1).

## 2. Materials and methods

### 2.1. Chemicals

A database containing 95 different surfactants (Cocoamide, Sodium Laureth Sulfate, Betaine ...) has been compiled (supplementary appendices) from 14 different suppliers (producers or resellers). This database will be used for the identification of surfactants contained in commercial detergents or homemade detergent mixtures.

### 2.2. Sample preparation

For sample preparation, there are two possible scenarios: either it involves a completely unknown mixture, or the constituents are known. In the case of an unknown mixture, the raw material will be diluted by a factor of 2, 3, etc. If the constituents are known, however, no sample preparation is required beforehand. The RAMAN spectrum of the commercial product or the house mixture will be analyzed. Identification will be performed from the RAMAN spectra databases of the commercial raw materials and quantification from the PM mixtures database.

### 2.3. Data base preparation

#### 2.3.1 Spectral database

A Raman spectral database is being created using a library of commercial raw materials. For each raw material, "pure" Raman spectra and diluted Raman spectra of the raw materials are recorded. The diluted spectra will be prepared at dilution levels of 75%, 50%, 25%, and 5%. This database has been constructed using 95 different commercial raw materials, resulting in a total of 380 Raman spectra. This comprehensive database will enable the identification of raw materials present in our various mixtures.

#### 2.3.2 Mixture database

A comprehensive database of commercial raw material mixtures is currently being acquired. These mixtures are composed of 2 to 5 components carefully selected and blended. To conduct in-depth investigations involving mixtures with 3, 4, and 5 components, it is imperative to prepare a minimum of 10, 18, or 30 mixtures respectively, following the highly effective Scheffé simplex designs strategy[25], [26]. It is worth noting that certain raw materials have specific constraints regarding their permissible usage concentrations, as specified in their corresponding safety data sheets. These constraints were meticulously considered during the formulation of the mixtures. The extensive research effort resulted in the preparation and analysis of over 1000 meticulously crafted mixtures, yielding valuable insights and data.

### 2.4. Measurement for RAMAN spectra of surfactants dishwashing product

Raman Rxn1 spectrometer (Kaiser Optical Systems, Inc. USA), equipped with a thermoelectrically cooled CCD detector, was used in combination with a fiber optic sapphire immersion probe. The laser wavelength was set at 785 nm. All spectra were recorded at a resolution of 4 cm$^{-1}$ in the spectral range from 150 to 3480 cm$^{-1}$. Acquisition time was set at 5 second and five spectra were accumulated.

## 2.5. Statistical analysis

### 2.5.1. Data preprocessing

To accentuate specific spectral variations, preprocessing of the spectra was conducted. Initially, the spectra were normalized to address any scale and baseline shift influences. To normalize and rectify noise, a multiplicative signal correction (MSC) method was employed[27]. MSC is a relatively straightforward preprocessing technique that aims to compensate for scaling and offset (baseline) effects. This correction was accomplished by regressing a measured spectrum against a reference spectrum and subsequently adjusting the measured spectrum based on the slope (and potentially intercept) of this regression. Each spectrum was corrected to achieve a comparable scatter level to that of the reference spectrum.

### 2.5.2. Independent Component Analysis (ICA)

ICA[17], [20]–[22], [28] is one of the most powerful techniques for blind source separation. ICA aims at identifying the products present in a mixture during a process. The basic assumption of the ICA is to consider each row of matrix X as a linear combination of "source" signals, S, with weighting coefficients, or "proportions", A, proportional to the contribution of the source signals in the corresponding mixtures. Its objective is to extract the "pure" components from a data set mixed in unknown proportions. For an unnoised model, the matrix X (s × n) is decomposed into f independent source signals S (f × n) and a mixing proportion matrix of these pure signals A (s × f) according to the following expression:

$$X = AS \qquad (1)$$

To solve equation ($A.1$), ICA estimates a unmixing matrix W (equal to $A^{-1}$) that optimizes the product independence between this matrix and the data matrix X according to an iterative method based on the central limit theorem [29] (which states that a sum of independent and identically distributed random variables tends to a Gaussian random variable). The output U must be as independent as possible. For a noise-free model, W must be the inverse of A and U must be equal to S, according to the following equation:

$$U = WX = W(AS) = S \qquad (2)$$

The unmixing matrix *A* can be calculated as:

$$A = XS^T(SS^T)^{-1} \qquad (3)$$

In this work, InfoMax[16], [22] implementation of the ICA algorithm was used. InfoMax uses Gram-Schmidt orthogonalization to ensure the independence of the extracted signal. It uses a maximum likelihood formulation. The aim of Infomax is to find independent source signals by maximizing entropy:

$$H(x) = -\int f(x) \log f(x)\, dx \qquad (4)$$

While independence of the signals cannot be measured, entropy can. Entropy is related to independence in that maximum entropy implies independent signals.

Therefore, the objective of ICA is to find the unmixing matrix that maximizes the entropy in the extracted signals.

### 2.5.3 Number of components

If too few ICs are extracted, some of the significant components may remain in the residual matrix; on the other hand, if to many ICs are extracted some of the significant components might themselves be decomposed into subcomponents. Validation methods are required to decide about the optimal number of ICs to be used in the computation of the final model. The ICA_by_blocks algorithm[8], [12], [30] was used to determine the optimal number of signals to extract. The initial data matrix is split into B blocks of samples with approximately equal numbers of rows. A ICA models are then computed with an increasing number of ICs for each block. The independent components calculated should be strongly correlated.

### 2.5.5 Model calibration

For each product to deformulate, the composition was determined through a calibration conducted using Partial Least Squares Regression (PLSR)[31]–[33]. PLSR is a commonly employed method, particularly when analyzing extensive spectral data. In essence, the algorithm for this regression is partially derived from the one used in Principal Component Analysis (PCA)[34], as it involves a dual decomposition into latent variables for both the X matrix of variables and the Y matrix of responses.

The development of the PLS model relies on the establishment of a mixing plan specifically designed for the species identified in the products to be reformulated. If we have access to the safety data sheet (SDS) of the said product, it will impose constraints on the constituents. Consequently, we will be able to adjust the mixing plan based on these constraints. In the absence of an SDS, a mixing plan comprising 2 to 5 components will be devised using the mixture database.

### 2.5.6 Software

Data collection was controlled using the HoloGRAMS™ software (Kaiser optical systems, Inc. USA). All spectra were imported into Matlab 9.1 (R2016b) (Mathworks, Natick, Massachusetts, USA). Statistical analyses were performed with the PLS-toolbox 8.2 (Eigenvector Research Incorporated Wenatchee, Washington, USA) and codes developed in-house.

## 2 Results and discussion

The method was tested on 8 dive products: 5 belonging to a commercial range whose composition was unknown, and 3 with constituents known through their safety data sheets. To verify the methodology, the constitution and composition of the unknown products were provided only at the end of the study.

For the identification of the constituents in the 5 unknown mixtures, dilutions were performed according to the described protocol, and Raman spectra were obtained. Only constituents present in the mixture to be analyzed at a concentration greater than 1% will be considered in the Independent Component Analysis (ICA). The ICA, using the ICA by block method (Figure.2), will determine the number of visible constituents per studied mixture and calculate a theoretical spectrum for each identified constituent. Among the extracted Independent Components (ICs), only those representing reliable information will be discussed. These spectra will be compared with the spectral library acquired through spectral overlay and correlation between calculated and experimental spectra (Figure.3). During the calculation of ICs, several similar spectra can be obtained. This can occur because certain constituents in the mixture may have similar spectra. In this case, it was decided to include surfactants with similar spectra in the algorithm within the mixture space. Detergents may contain additional compounds that can be detected by Raman spectroscopy, such as salts ($NaCl$, $MgSO_4$, etc.). These salts are usually added to increase the viscosity of the mixture. These salts have specific bands in Raman spectroscopy, such as the vibration band of $SO_4^{2-}$ at 2550 cm$^{-1}$. The addition of salt is considered when constructing the mixture plan, as its presence may interact with certain acidic or basic surfactants.

Next, a mixture plan is constructed, either using the spectral profiles from the library's mixture plans or by performing new mixtures, considering the specificities identified during the ICs calculation. Whenever a specific blending plan is required, typically due to specific constraints on the components, that plan is systematically added to the blend database. The points in the mixture plan will serve as calibration points to establish a model that allows us to determine the composition of the mixture.

In the case of the 3 products with known constituents, a mixture plan is established based on the database of mixtures, while respecting the constraints described in the raw material safety data sheets (SDS). The Partial Least Squares (PLS) modeling is then used to determine the composition of the studied mixture. For both approaches, to validate the methodology, the criteria for prediction errors (Root Mean Square Error of Calibration (RMSEC), Root Mean Square Error of Cross-Validation (RMSECV), and Root Mean Square Error of Prediction (RMSEP)) are observed, as well as the determination coefficient R² for calibration, cross-validation, and prediction[35]. The statistical criteria for the models involving unknown mixtures (*5) and known mixtures (*3) are presented in Tables 1. Based on these criteria, the prediction and calibration residuals are of similar magnitudes, indicating a good predictive quality for the mixture compositions. For all the model's prediction results, the obtained compositions fall within the confidence interval of the provided compositions (Table 2 and 3). In both cases, the results generally demonstrate accurate estimation of the constituents in the various mixtures. The prediction discrepancies, although minimal for quantifying an estimation of the mixture compositions, can have various origins, such as the nature of a raw material, interactions with co-constituents in the mixture (such as fragrances, thickeners, etc.), or the concentration of a constituent. Regarding the nature of the raw material, even for the same constituent, there are numerous producers and suppliers, and often there exist slight differences between these constituents, such as variations in carbon chain length or the number of ethoxylated groups, which can impact the spectrum and therefore its prediction. A constituent present in low concentration would be difficult to detect in Raman spectroscopy and consequently be identified, as is the case with MGDA in samples D1 and D3. Hence, the model would have a higher prediction error in cases of low concentration.

## 4      Conclusions

Chemometrics/Machine Learning methods such as Blind Source Separation (BSS) are powerful tools for extracting signals from complex mixtures. These techniques have been successfully applied to several detergent mixtures for various household applications. The combination of spectral databases of surfactants and mixtures has enabled the identification and quantification of surfactants in these complex mixtures (Fig.3).

This methodology can be easily adapted to industrial environments to perform various tasks such as raw material quality control and competitive intelligence monitoring. The methodology can be applied to any type of MIR, NIR, and Raman spectroscopy. Of course, it is necessary to redo all the measurements to obtain the various databases required for the identification and quantification of the mixture.

This approach could facilitate rapid monitoring of detergent type and concentration in different matrices. This analysis would make it possible to determine which types of detergents are present, as well as their respective concentrations. This information could then be used to adjust methods to better eliminate or reduce the specific detergents detected.

**AUTHOR INFORMATION**

**Corresponding Author**
Yohann Clément – Data Scientist / Chemometrician, University of Lyon, CNRS, Institut of Analytical Sciences, UMR-5280, 5 Rue de la Doua 69100 Villeurbanne, France; orcid.org/0000-0002-9852-2856; Email: yohann.clement@univ-lyon1.fr

**Author**
Pedro Marote, Analyst, University of Lyon, CNRS, Institut of Analytical Sciences, UMR-5280, 5 Rue de la Doua 69100 Villeurbanne, France; Email: pedro.marote@univ-lyon1.fr

Pierre Lanteri, Professor, University of Lyon, CNRS, Institut of Analytical Sciences, UMR-5280, 5 Rue de la Doua 69100 Villeurbanne, France ; orcid.org/ 0000-0002-8244-9834; Email: pierre.lanteri@univ-lyon1.fr

Marie Martin, Professor, University of Lyon, CNRS, Institut of Analytical Sciences, UMR-5280, 5 Rue de la Doua 69100 Villeurbanne, France ; Email: marie.martin@isa-lyon.fr

Anne Bonhommé, Professor, University of Lyon, CNRS, IRCELYON, UMR-5256, 5 Rue de la Doua 69100 Villeurbanne, France; Email: anne.bonhomme@ircelyon.univ-lyon1.fr

**Author Contributions**
The manuscript was written through contributions of all authors. All authors have given approval to the final version of the manuscript.

**Funding Sources**

This research did not receive any specific grant from funding agencies in the public, commercial, or not-for-profit sectors.

**Declarations of interest**
None.

**Acknowledgements**
None**ABBREVIATIONS**
NMR: Nuclear Magnetic Resonance, BSS: Blind source separation, SE: Source extraction, ICA: Independent Component Analysis, MCR-ALS: Multicurve Resolution Alternating Least Squares, ICs: Independent Components, MSC: multiplicative signal correction, PLSR: Partial Least Squares Regression, PCA: Principal Component Analysis, SDS: safety data sheet, RMSE: Root Mean Square Error, RMSEC: Root Mean Square Error of Calibration, RMSEP: Root Mean Square Error of prediction, RMSECV: Root Mean Square Error of Cross-Validation

| INCI | RMSEC | RMSECV | RMSEP | $R^2Y$ | $Q^2Y$ |
|---|---|---|---|---|---|
| Sodium C14-16 Olefin Sulfonate | 0.93 | 1.2 | 1.22 | 0.98 | 0.97 |
| Sodium Laureth Sulfate | 0.48 | 1 | 0.87 | 0.98 | 0.94 |
| Trimethyl Amine (TEA) | 0.27 | 0.33 | 0.39 | 0.99 | 0.98 |
| Trisodium salt of Methylglycinediacetic acid (MGDA) | 0.61 | 1.6 | 0.68 | 0.94 | 0.86 |
| Lauryl ether sulfate | 0.76 | 0.92 | 0.98 | 0.99 | 0.98 |
| Eau | 1.78 | 2.15 | 2.86 | 0.98 | 0.99 |

Table 1: RMSEC, RMSECV, RMSEP, $R^2Y$ and $Q^2Y$ for PLS regression on raw material detected by Independent Component Analysis (ICA)

| Unknowm Sample | Sodium C14-16 Olefin Sulfonate | | Sodium Laureth Sulfate | | Trimethyl Amine (TEA) | | Trisodium salt of Methylglycinediacetic acid (MGDA) | | Lauryl ether sulfate | |
|---|---|---|---|---|---|---|---|---|---|---|
| | Experimental | calculated | Experimental | calculated | Experimental | calculated | Experimental | calculated | Experimental | calculated |
| D1 | 8.4 | 8.8 | 4.5 | 5.0 | 6.1 | 5.8 | 1.2 | 0.8 | 39.7 | 37.6 |
| D2 | 7.4 | 7.2 | 8.8 | 9.5 | 9.9 | 9.4 | 4.3 | 3.6 | 29.0 | 24.4 |
| D3 | 12.0 | 13.2 | 7.2 | 6.3 | 4.3 | 4.6 | 0.9 | 1.8 | 0.0 | 3.1 |
| D4 | 14.5 | 15.3 | 9.8 | 7.1 | 6.1 | 5.5 | 2.7 | 3.5 | 0.0 | 2.1 |
| D5 | 16.8 | 16.3 | 12.7 | 11.5 | 7.5 | 8.2 | 4.4 | 5.1 | 0.0 | 0.8 |

Table 2: Composition of unknown detergent: experimental vs calculated by PLS regression.

| Known sample | Sodium C14-16 Olefin Sulfonate | | Sodium Laureth Sulfate | | Cocamdopropyl betaine | | Lauryl ether sulfate | | Lauramidopropylamine Oxide | |
|---|---|---|---|---|---|---|---|---|---|---|
| | FDS | calculated | Experimental | calculated | Experimental | calculated | Experimental | calculated | Experimental | calculated |
| PC | 0% | 0% | 10 15 % | 14% | 0% | 0% | 0% | 0% | 5-10 % | 8.5% |
| RA | 5-10 % | 8% | < 1% | 0.50% | 1-5 % | 2.7% | 1-5 % | 4% | 0% | 0% |
| MI | 0% | 0% | 5-10 % | 9% | 1-5 % | 3.5% | 0% | 0% | 0% | 0% |

Table 3: Composition of known detergent by FDS: experimental vs calculated by PLS regression.

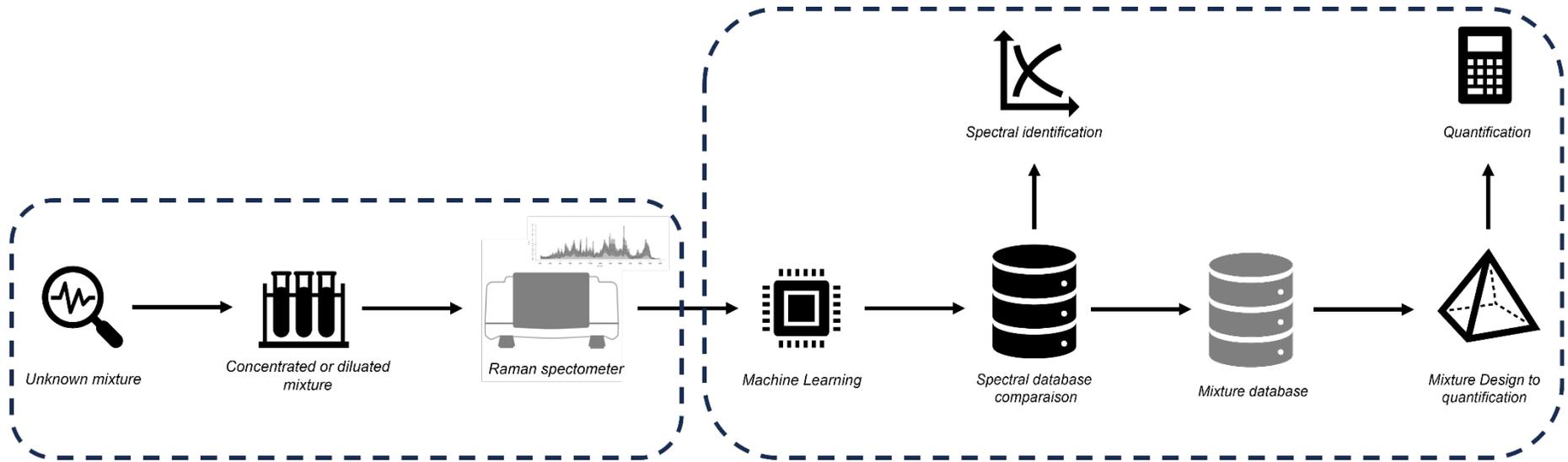
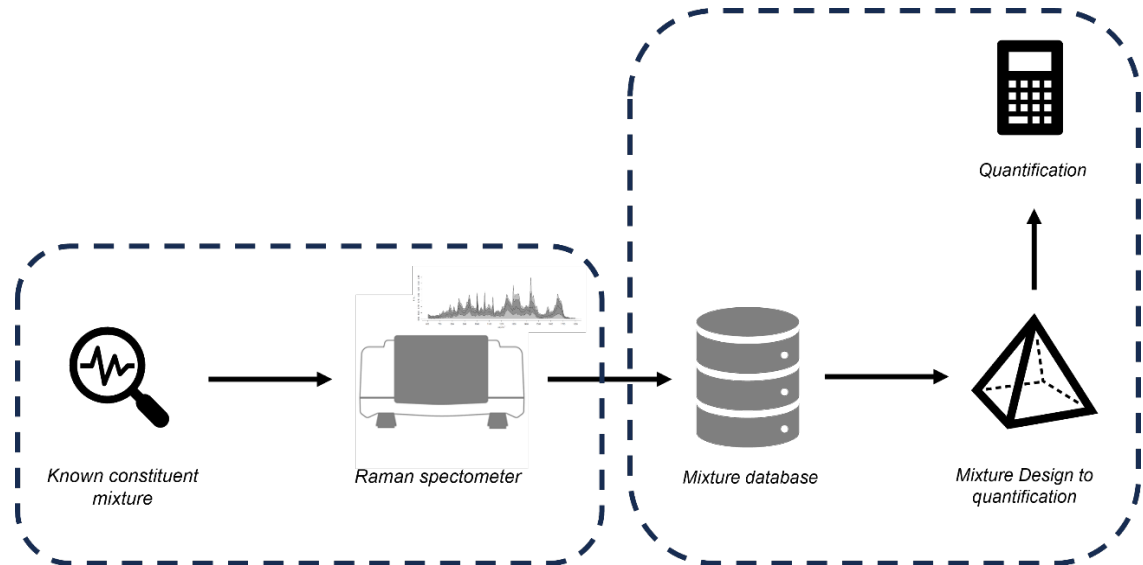

Figure.1: to Reverse Engineering of Detergents Using Raman Spectroscopy

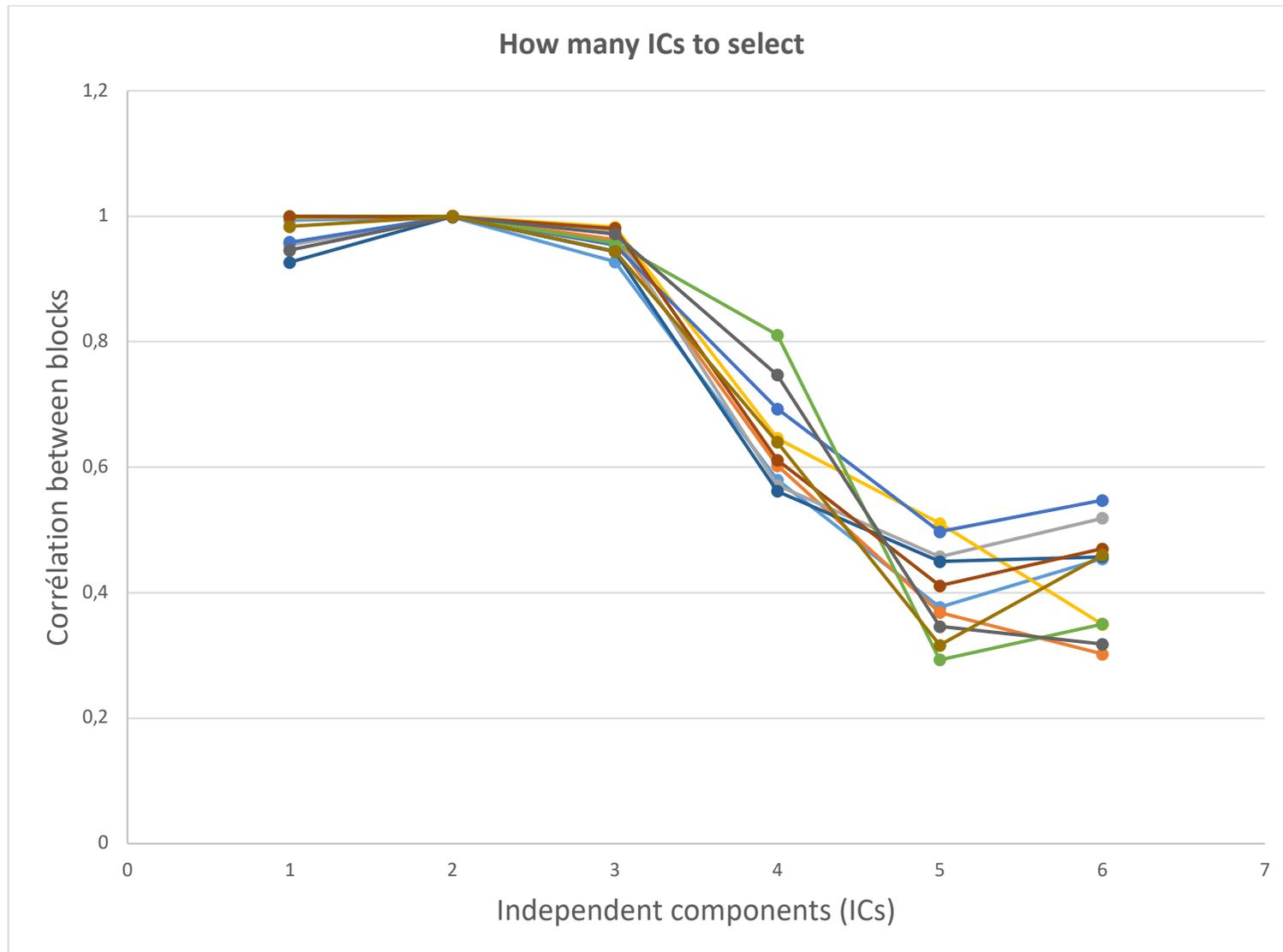

11
12 Figure.2 ICA by blocks tests for the determination of raw number materials in the detergent.

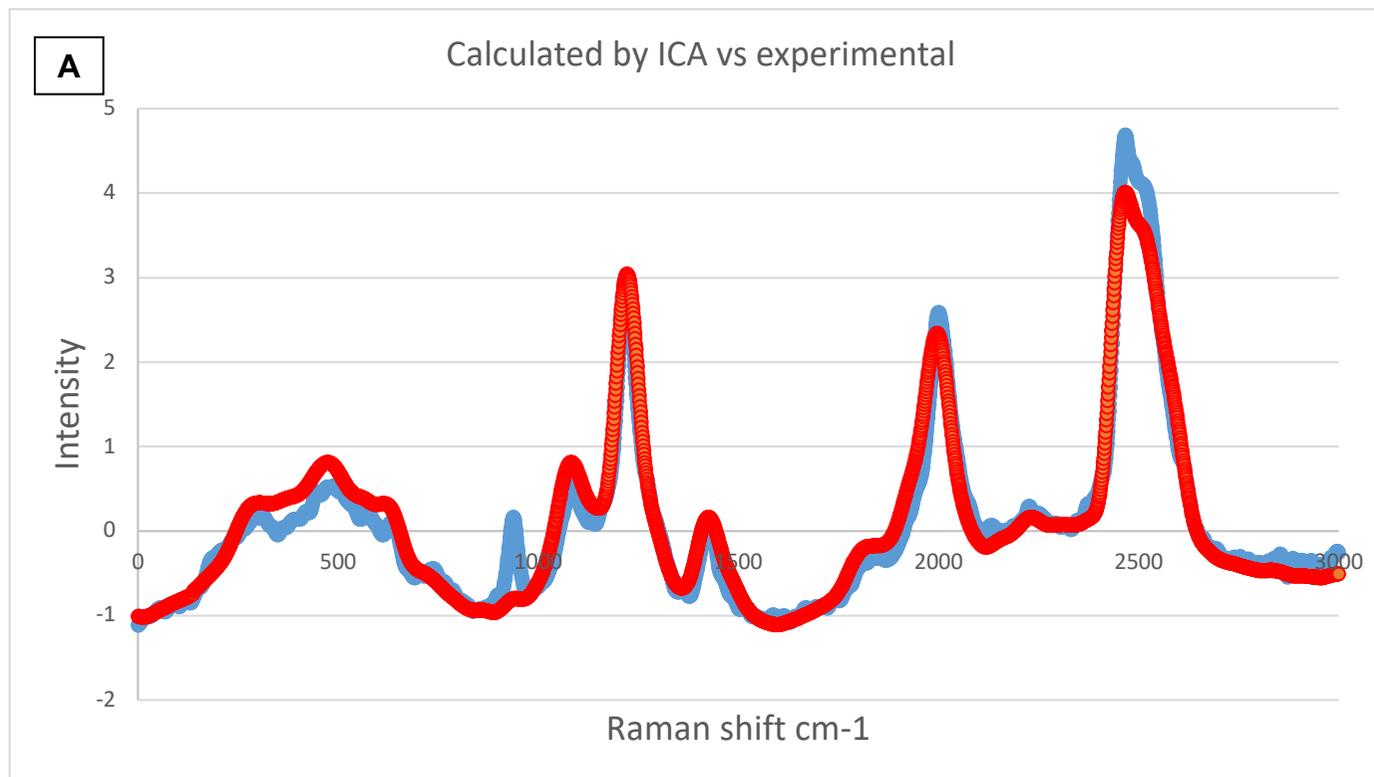

14
15
16

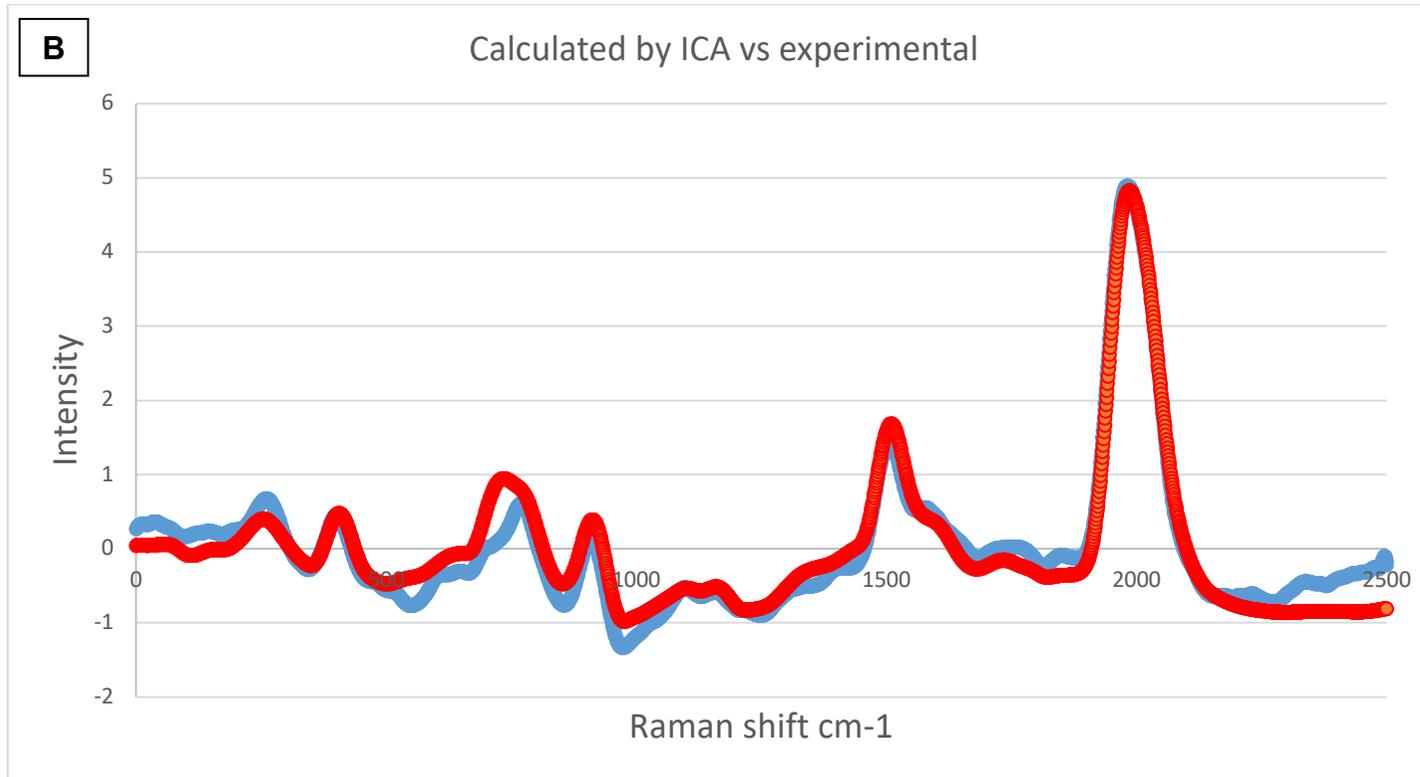

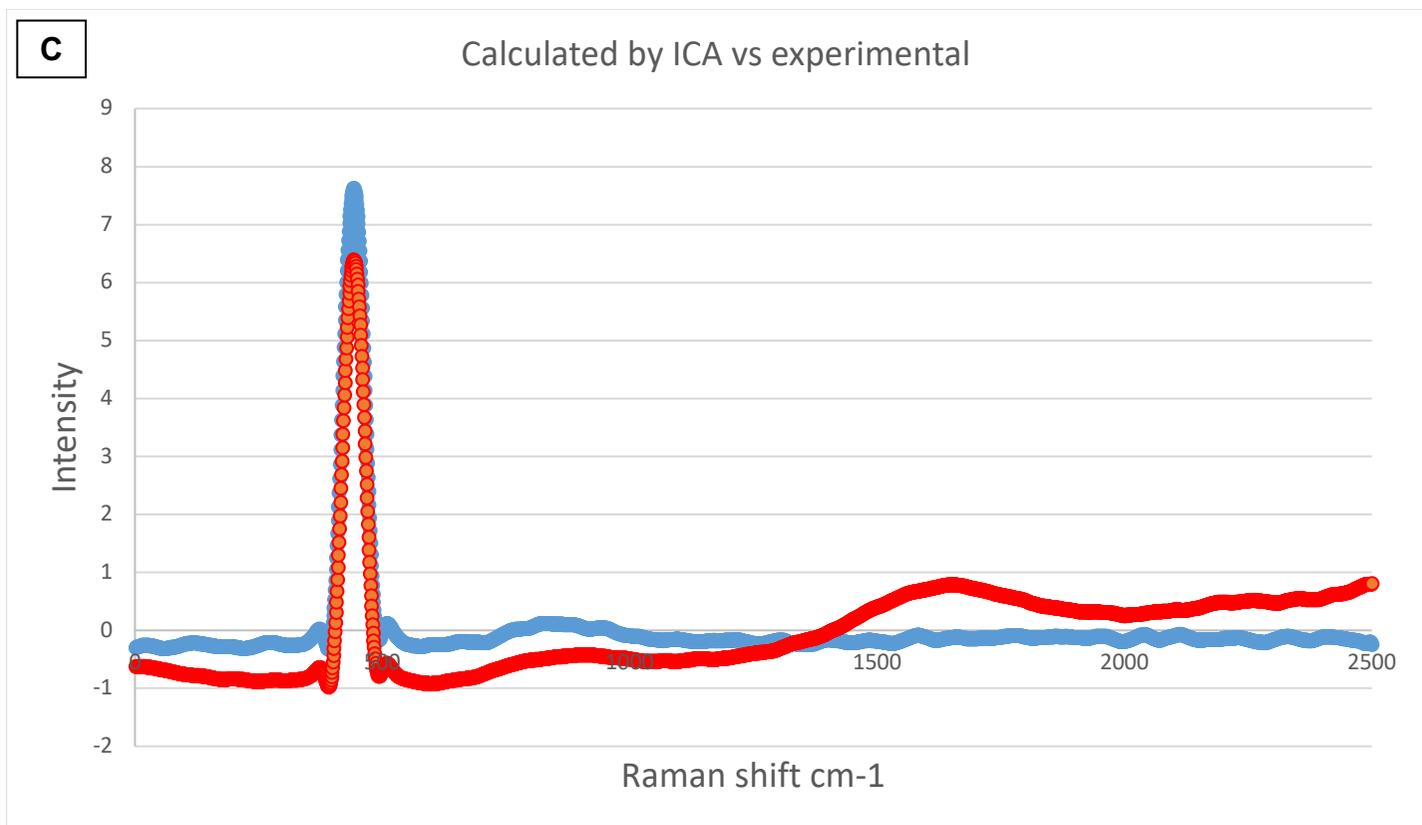

Figure. 3. Raman spectra of 2 main raw materials (A and B) and 1 salt (C) (blue) versus calculated spectra according ICA (red) A) Sodium Laureth Sulfate, B) Lauramidopropylamine Oxide and C) $SO_4^{2-}$ .

32
33 **Supplementary appendices**

| Name | INCI name | Producer / Reseller |
|---|---|---|
| Dehyquart ECA | 1-Hexadecanaminium, N,N,N-trimethyl-, chloride | BASF |
| Dhyton K Cos | 1-Propanaminium, 3-amino-N-(carboxymethyl)-N,N-dimethyl-, N-(C8-18 and C18-unsaturated acyl) derivatives, hydroxides, inner salts | BASF |
| Polyquart H81 | 1,3-Propanediamine, N-(3-aminopropyl)- | BASF |
| Luviquat Excellence | 1H-Imidazolium, 1-ethenyl-3-methyl-, chloride, polymer with 1-ethenyl-2-pyrrolidinone | BASF |
| Lanette O | Alcohols, C16-18 | BASF |
| Emulgin B2 | Alcohols, C16-18, ethoxylated | BASF |
| Comperlan 100 | Amides, C12-18 and C18-unsaturated, N-(hydroxyethyl) | Ami |
| Comperlan IP | Amides, coco, N-(2-hydroxypropyl) | Ami |
| Stepanol AM 30 KE | Ammonium lauryl sulfate | Stepan |
| Betafin BP 20 | Betaine (anhydre 99%) | Masso |
| Dehyton AB 30 | Betaines, C12-14-alkyldimethyl | BASF |
| Cosmacol ELI | C12-13 Alkyl Lactate | Sasol |
| Dehyquart F75T | Ceteareth-20 | Ami |
| Emilgin B2 | ceteareth-20 | Cognis |
| Dehyquart ACA | Cetrimonium Chloride | Ladybel |
| Amphosol CDB special | Cetyl Betaine | Stepan |
| Hydrogen CAT | cetyl PEG/PPG-10/1 dimethicone | Cognis |
| Ninol 40 CO E | Cocamide DEA | Stepan |
| Purton CFD | COCAMIDE DEA | ZW |
| Comperlan 100 | cocamide MEA | Cognis |
| Purton CFM/ F | Cocamide MEA | ZW |
| Comperlan IP | Cocamide MIPA | Cognis |
| Emulgin B2 | Cocamide MIPA | Ami |

| Trade Name | INCI | Supplier |
|---|---|---|
| Amphotensil B4/C | Cocamidopropyl Betaine | ZW |
| Amphosol DM | Cocamidopropyl Betaine | Stepan |
| Amphotensid B5 | Cocamidopropyl Betaine | ZW |
| Tegobetaine F 50 | cocamidopropyl betaine | Cognis |
| Amphosol CG-K | Cocamidopropyl Betaine | Stepan |
| Antil HS 60 | cocamidopropyl betaine ; glyceryl laurate | Cognis |
| Eco sense 919 surfactant | Coco-Glucoside | Dow |
| Plantacare 818 UP | COCO-GLUCOSIDE | BASF |
| Liviquat mono LS | Cocotrimonium methosulfate | BASF |
| Plantacare 2000UP | Decyl Glucoside | BASF |
| Ninol CCA | Dimethyl lauramide | Stepan |
| Texapon N40 IS | Disodium 2-Sulfolaurate | BASF |
| Miranol | Disodium Cocoamphodiacetate | Rhone Poulenc |
| Setacin 103 Spezal | Disodium Laureth Sulfosuccinate | ZW |
| Stepan MILD SL3 BA | Disodium Laureth Sulfosuccinate | Stepan |
| Rewopol SB CS50K | disodium PEG-5 laurylcitrate sulfosuccinate ; sodium laureth sulfate | Cognis |
| Dehyquart F75T | Distearoylethyl Hydroxyethylmonium Methosulfate (and) Cetearyl Alcohol | BASF |
| Trilon B 87% | EDTA | BASF |
| Stepan MILD GCC | Glyceryl Caprylate/Caprate | Stepan |
| Tegin BL 315 | glycol destearate | Cognis |
| dehyquart N | Guar gum, 2-hydroxy-3-(trimethylammonio)propyl ether, chloride | BASF |
| Tegosoft P | isopropyl palmitate | Cognis |
| ammonyx LMDO | Lauramidopropylamine Oxide | Stepan |
| Ammonyx LO | Lauramine Oxide | Stepan |
| Empigen OB | Lauramine Oxide | Innospec Performance Chemicals |

| Product | INCI/Chemical Name | Supplier |
|---|---|---|
| Plantacare 1200UP | Lauryl Glucoside | BASF |
| Stepan MILD L3 | LAURYL LACTYL LACTATE | Stepan |
| Abilsoft AF100 | methoxy PEG/PPG-7/3 aminopropyl dimethicone | Cognis |
| Zetesol 2056 | MIPA-Laureth Sulfate | ZW |
| Lumorol K 1056 | MIPA-Laureth Sulfate, Cocamidopropyl Betaine | ZW |
| Arlypon LIS | Oxirane, 2-methyl-, polymer with oxirane, ether with 2-ethyl-2-(hydroxymethyl)-1,3-propanediol (3:1), tri-(9Z)-9-octadecenoate | BASF |
| Arlypon LIS | Oxirane, 2-methyl-, polymer with oxirane, ether with 2-ethyl-2-(hydroxymethyl)-1,3-propanediol (3:1), tri-(9Z)-9-octadecenoate | Ladybel |
| Arlypon TT | Oxirane, 2-methyl-, polymer with oxirane, ether with 2-ethyl-2-(hydroxymethyl)-1,3-propanediol (3:1), tri-(9Z)-9-octadecenoate | Ami |
| Arlypon TT | Oxirane, 2-methyl-, polymer with oxirane, ether with 2-ethyl-2-(hydroxymethyl)-1,3-propanediol (3:1), tri-(9Z)-9-octadecenoate | BASF |
| Antil 171 | PEG-18 glyceryl oleate/cocoate | Cognis |
| Rewoderm LIS 80 | PEG-200 hydrogenated grylceryl palmate (and) PEG-7 glyceryl cocoate | Cognis |
| Arlacel P 135 | PEG-30 Dipolyhydroxystearate | Masso |
| Arlyton TT | PEG/PPG-120/10 trimethylolpropane trioleate (and) laureth-2 | Cognis |
| Myritol 318 | PEG/PPG-120/10 Trimethylolpropane Trioleate (and) Laureth-2 | Ami |
| Texapon SB 3KC | Poly(oxy-1,2-ethanediyl), .alpha.-(3-carboxy-1-oxosulfopropyl)-.omega.-hydroxy-, C10-16-alkyl ethers, disodium salts | Ami |
| Isolan GO 3 | polyglyceril 3 oleate | Cognis |
| Emulgin S21 | Polyoxyethylene monooctadecyl ether | BASF |
| Emulgin S21 | Polyoxyethylene monooctadecyl ether, C18H37O(C2H4O)21H | Ami |
| Salcare SL92 | polyquaternium-32 (and) mineral oil (and) PPG-1 trideceth-6 | BASF |
| Polysorbate 20 | Polysorbate 20 | Ladybel |
| Tween 21 LQ | Polysorbate 21 | Masso |
| Tween 60V | Polysorbate 60 | Masso |
| Amphisol K | Potassium Cetyl Phosphate | DMS |

| Name | Composition | Supplier |
|---|---|---|
| Bio Terge AS 40 HASB | Sodium C14-16 Olefin Sulfonate | Stepan |
| Dehyton MC | Sodium cocoamphoacetate | Ladybel |
| Rowoteric AMC | sodium cocoamphoacetate | Cognis |
| Chimin CG | SODIUM COCOYL GLUTAMATE | Lamberti |
| Protelan GG | Sodium Cocoyl Glycinate, Sodium Cocoyl Glutamate | ZW |
| Steol CS 270 | Sodium Laureth Sulfate | Stepan |
| Zetesol | Sodium Laureth Sulfate | ZS |
| Zetesol 370 /N | Sodium Laureth Sulfate | ZW |
| Zetesol LES 2 | Sodium laureth sulfate | ZW |
| Zetesol NL U | Sodium Laureth Sulfate | ZW |
| Steol 370 | Sodium Laureth Sulfate | Stepan |
| Perlagent GM 4175 | Sodium Laureth Sulfate, Glycol Stearate, Cocamide MEA, Cocamide DEA, Propylene Glycol | ZW |
| Lumorol K 5240 | Sodium Laureth Sulfate, Cocamido- propyl Betaine, Disodium Laureth Sulfosuccinate, PEG-9 Cocoglycerides | ZW |
| Miranol ultra L32 E | Sodium lauroamphoacetate | Solvay |
| Maprosil 30B | Sodium Lauroyl Sarcosinate | Stepan |
| Protelan LS 9011 | Sodium Lauroyl Sarcosinate | ZW |
| Sulfetal LS U | Sodium Lauryl Sulfate | ZW |
| SDS | Sodium Lauryl Sulfate | Aldrich |
| Lathanol LAL coarse | Sodium Lauryl Sulfoacetate | Stepan |
| Stepanate SXS E | SODIUM XYLENE SULFONATE | Stepan |
| Purton SFD | SOYAMIDE DEA | ZW |
| Copherol 1300C | tocopherol | Cognis |
| EMPILAN 2502 | coconut diethanolamide | Innospec Performance Chemicals |
| TRILON M | Trisodium salt of Methylglycinediacetic acid (MGDA) | BASF |

34  Table supplementary: raw material list with the commercial name, the INCI name and the producer or the reseller
35
36